\documentclass[letterpaper, 10 pt, journal, twoside]{IEEEtran}  %

\IEEEoverridecommandlockouts                              

\usepackage{graphicx} 
\usepackage{amsmath} 
\usepackage{amssymb}  
\usepackage[usenames, dvipsnames]{color}

\usepackage{lipsum}
\usepackage{color}
\usepackage{cite}

\usepackage{diagbox}
\usepackage{tabu}
\usepackage{balance}  

\usepackage[ruled,linesnumbered]{algorithm2e}
\usepackage{algpseudocode}
\usepackage{epsfig}
\usepackage{multirow}

\usepackage{enumitem}
\usepackage{setspace}
\usepackage{tikz}
\usepackage{svg}
\usepackage{makecell}

\usepackage{flushend}

\newcommand{\revision}[1]{\textcolor{black}{#1}} 

\usepackage{hyperref}

\title{
Needle Segmentation Using GAN: Restoring Thin Instrument Visibility in Robotic Ultrasound  

}

\author{Zhongliang Jiang*, Xuesong Li*, Xiangyu Chu, Angelos Karlas, \\ Yuan Bi, Yingsheng Cheng, K. W. Samuel Au, and Nassir Navab, \textit{Fellow, IEEE} 
\thanks{$^{*}$ Authors with equal contributions.}
\thanks{Z. Jiang, X. Li, Y. Bi and N. Navab are with the Chair for Computer Aided Medical Procedures and Augmented Reality (CAMP), Technical University of Munich, 85748 Garching, Germany. {\tt\footnotesize{(Corresponding authors: Yingsheng Cheng and Xiangyu Chu)}}}

\thanks{A. Karlas is with the Department for Vascular and Endovascular Surgery, Klinikum rechts der Isar, Technical University of Munich, Munich, Germany and also with DZHK (German Centre for Cardiovascular Research), Munich, Germany.}

\thanks{Y. Cheng is with the Department of Imaging Medicine and Nuclear Medicine, Shanghai Tongji Hospital, Tongji University School of Medicine.  }%

\thanks{X. Chu and K.W. Samuel Au are with the Department of Mechanical and Automation Engineering, The Chinese University of Hong Kong, Hong Kong SAR, China and Multi-scale Medical Robotics Centre, Hong Kong SAR, China.    }%
}

\begin{document}

\maketitle


\begin{abstract}
Ultrasound-guided percutaneous needle insertion is a standard procedure employed in both biopsy and ablation in clinical practices. However, due to the complex interaction between tissue and instrument, the needle may deviate from the in-plane view, resulting in a lack of close monitoring of the percutaneous needle. To address this challenge, we introduce a robot-assisted \revision{ultrasound (US)} imaging system designed to seamlessly monitor the insertion process and autonomously restore the visibility of the inserted instrument when misalignment happens. To this end, \revision{the adversarial structure is presented to encourage the generation of segmentation masks that align consistently with the ground truth in high-order space.} 
This study also systematically investigates the effects on segmentation performance by exploring various training loss functions and their combinations. When misalignment between the probe and the percutaneous needle is detected, \revision{the robot is triggered to perform transverse searching to optimize the positional and rotational adjustment to restore needle visibility.} The experimental results on ex-vivo porcine samples demonstrate that the proposed method can precisely segment the percutaneous needle (with a tip error of $0.37\pm0.29mm$ and an angle error of $1.19\pm 0.29^{\circ}$). Furthermore, the needle appearance can be successfully restored under the repositioned probe pose in all 45 trials, with repositioning errors of $1.51\pm0.95mm$ and $1.25\pm0.79^{\circ}$. 
\end{abstract}




\bstctlcite{IEEEexample:BSTcontrol}

\section{Introduction}

\begin{figure}[ht!]
\centering
\includegraphics[width=0.48\textwidth]{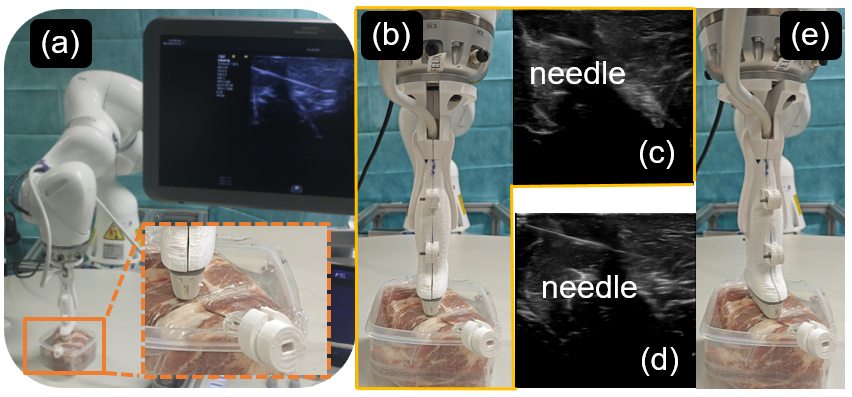}
\caption{Illustration of needle insertion on an ex vivo sample of porcine tissue. (a) is the overall scene. (b) and (e) refer to the misalignment case and the desired case, respectively. (c) and (d) are the corresponding B-mode images obtained in the case of (b) and (e), respectively.
}
\label{Fig_demo_question}
\end{figure}

\IEEEPARstart{P}{recise} needle placement is a crucial procedure in many minimally invasive interventions, such as tissue biopsy and cancerous tissue ablation. Given its radiation-free nature and real-time capabilities, medical ultrasound (US) has been widely used in providing live images to guide needle insertion, thereby enhancing targeting accuracy~\cite{masoumi2023big, yang2023medical}. However, it is challenging to maintain the needle inside the traditional two-dimensional (2D) US plane during the needle insertion. Certain training and effort are needed for clinical experts to coordinate the alignment between the US probe and the inserted needle, particularly considering US images often suffer from speckle noise, artifacts, and low contrast. Therefore, precise needle segmentation and tracking is a vital and essential step toward accurate needle placement for both biopsy and ablation therapy, particularly for robotic US systems (RUSS)~\cite{jiang2022precise, ottacher2020positional, yang2021automatic, li2021overview, jiang2023intelligent, li2024fully, huang2024robot, huang2023mimicking, bi2024machine, jiang2022towards}.

\par
The methods for segmenting instruments from US images can be broadly classified into external tracking methods and image-based methods. The former ones require additional devices, such as optical or electromagnetic tracking systems~\cite{zhao2019electromagnetic}. The latter methods are more convenient to be integrated into clinical practices. In this track, some traditional computer vision and machine learning technologies like Principal Component Analysis~\cite{novotny2003tool}, and Hough transform~\cite{beigi2021enhancement} have been tried. Besides, Kaya~\emph{et al.}~\cite{kaya2014needle} employed the Gabor filter to extract the instrument in US images and then used \revision{Random Sample Consensus (RANSAC)} line estimator to refine the continuity of the extracted shaft. Although decent results were reported in their scenarios, the performance of these conventional methods often suffers from inter and intra-patient property variations.

\par
Recently, Convolutional neural networks (CNNs) have been considered a promising alternative for tackling needle segmentation in US images. The rapid advancement of deep learning has led to the widespread application of CNNs in medical image segmentation~\cite{litjens2017survey}. For US image segmentation, Jiang~\emph{et al.} used classification results to improve US cartilage bone segmentation~\cite{jiang2024class}. To enhance the generalization capability of segmentation models, Bi~\emph{et al.} proposed a method that explicitly disentangles anatomical and domain features by calculating mutual information in the latent space~\cite{bi2023mi}. Specific to the task of instrument segmentation in US images, Gillies~\emph{et al.}~\cite{gillies2020deep} employed a U-Net architecture~\cite{ronneberger2015u} for the images of different anatomies, such as prostate and kidney. To take advantage of consecutive images, Chen~\emph{et al.} presented a W-Net architecture, where two images are fed to the two encoders simultaneously, while a single decoder is used to predict the inserted needle~\cite{chen2022automatic}. To enhance the prediction accuracy, Lee~\emph{et al.}~\cite{lee2020ultrasound} integrated spatial and channel “Squeeze and Excitation” (scSE) module into LinkNet~\cite{chaurasia2017linknet} to boost relevant features and suppressing less important ones in the intermediate feature space. To provide a cost-effective needle segmentation, Mwikirize~\emph{et al.}~\cite{mwikirize2018convolution} used a fast region-based CNN to extract anchor boxes. Subsequently, a positive label was assigned at the center of the predicted boxes. The tip was further extracted by searching along the extracted needle shaft as in~\cite{hacihaliloglu2015projection}. For accurate localization of the needle tip, Mwikirize~\emph{et al.}~\cite{mwikirize2021time} extended their approach by explicitly enhancing the needle tip feature based on consecutive US images. This enhanced tip map, along with B-mode images, was then combined and used as input for a segmentation network developed with CNN and Long Short-Term Memory (LSTM) modules.

\par
Beigi~\emph{et al.} extracted the micro-motion caused by needle insertion using spatiotemporal and spectral feature selections, therefore enabling needle segmentation and tracking when the needle appearance is imperceptible to the naked eye~\cite{beigi2017casper}. Likewise, to account for potential needle tip disappearance and background noise, Yan~\emph{et al.} introduced a learning-based needle tip tracking system~\cite{yan2023learning}. This system incorporates both a visual tracking module and a transformer-based motion prediction module to dynamically update the tip appearance in real-time and predict the current tip location, respectively. The aforementioned works mainly focused on the case that only has one needle. To tackle the challenge of multi-needle detection, Zhang~\emph{et al.} employed a self-supervised method to compute the residual images by using auxiliary images without needles~\cite{zhang2020multi}. 

\par
Recently, there has been growing interest in 3D US imaging due to the significant advantages of native 3D volumes, particularly in reducing the requirement for precise alignment of the needle with the US imaging plane. To segment the instrument in 3D US images, Arif~\emph{et al.} applied a modified 3D U-Net on liver US images and visualized the detected needle in two perpendicular cross-sectional planes from the 3D liver volume~\cite{arif2019automatic}. The reported position and orientation errors were $1~mm$ and $2^{\circ}$. To improve the time efficiency, Yang~\emph{et al.}~\cite{yang2020efficient} proposed a network using a 3D encoder to extract discriminating features and then use a 2D decoder to segment the instrument from the projected feature map in 2D space. Their results demonstrated the method can achieve a detection error of three voxels in $0.12~s$ per 3D input volume. Besides, Yang~\emph{et al.}~\cite{yang2021efficient} presented a POI-FuseNet with two main modules: 1) patch-of-interest (POI) selector to extract the region containing the instrument, and 2) FuseNet to fully exploit the contextual information by using both 2D and 3D \revision{Fully Convolutional Network (FCN)} features. However, native 2D array US transducers are still not widely used in clinical practices due to the limitation of time efficiency and restricted imaging volumes. So, this study works on the problem of 2D needle segmentation and tracking. It has been reported that it is challenging to maintain the needle in the in-plane view during 2D US-guided interventions. In clinical scenarios such as kidney or liver drainage procedures, clinicians need to ensure the needle's visibility in the plane to avoid the risk of accidental damage to adjacent organs.

\par
To address this challenge, this study presents a robotic repositioning system with the ability to track the inserted needle and autonomously reposition the US probe to recover the needle visibility once misalignment happens. The misalignment detection module is developed based on real-time segmentation. Upon detecting misalignment, the RUSS will autonomously reposition the probe to re-establish the tracking of the inserted needle. \revision{The main contributions are summarized as follows:
\begin{itemize}
  \item To precisely segment the intact needle in real-time, we propose an AdvSeg-Net by incorporating the adversarial component to encourage feature consistency and continuity in deep layers, aligning them with the ground truth in an unsupervised way.
  \item To deal with the imperfect annotation of thin needles, we systematically investigate the effects on final segmentation performance by using various training loss combinations and diverse network architectures as generators.
  \item To have a precise line-like representation of the needle shaft, this study first considers the importance of slice thickness artifact correction for images obtained when the needle is not aligned with the US view. 
\end{itemize}
}
Finally, to validate the effectiveness of the method, both segmentation results and probe repositioning performance were tested on ex vivo porcine tissue samples. The results demonstrated that the needle segmentation performance can reach $0.37\pm0.29~mm$ and $1.19\pm 0.29^{\circ}$ for tip error and angle error. The repositioning error are $1.51\pm0.95~mm$ and $1.25\pm0.79^{\circ}$. The \textbf{code}\footnote{Code: https://github.com/noseefood/NeedleSegmentation-GAN} and \textbf{video}\footnote{Video: https://youtu.be/4WuEP9PACs0} can be publicly accessed.


\par
The rest of this paper is organized as follows. Section II introduces the network architecture and the hybrid loss functions. Section III describes the components, including misalignment detection, slice thickness correction, and robotic probe repositioning computation. The experimental results on unseen porcine tissue samples are provided in Section IV. Finally, the summary of this study is presented in Section V. 

\section{Needle Segmentation in Ultrasound Images}   ~\label{sec:network_architecture}
In this section, we present the overall structure of the proposed AdvSeg-Net. In addition, various loss function combinations and the training process are described. The detailed description can be found in the following subsections.

\begin{figure*}[ht!]
\centering
\includegraphics[width=0.75\textwidth]{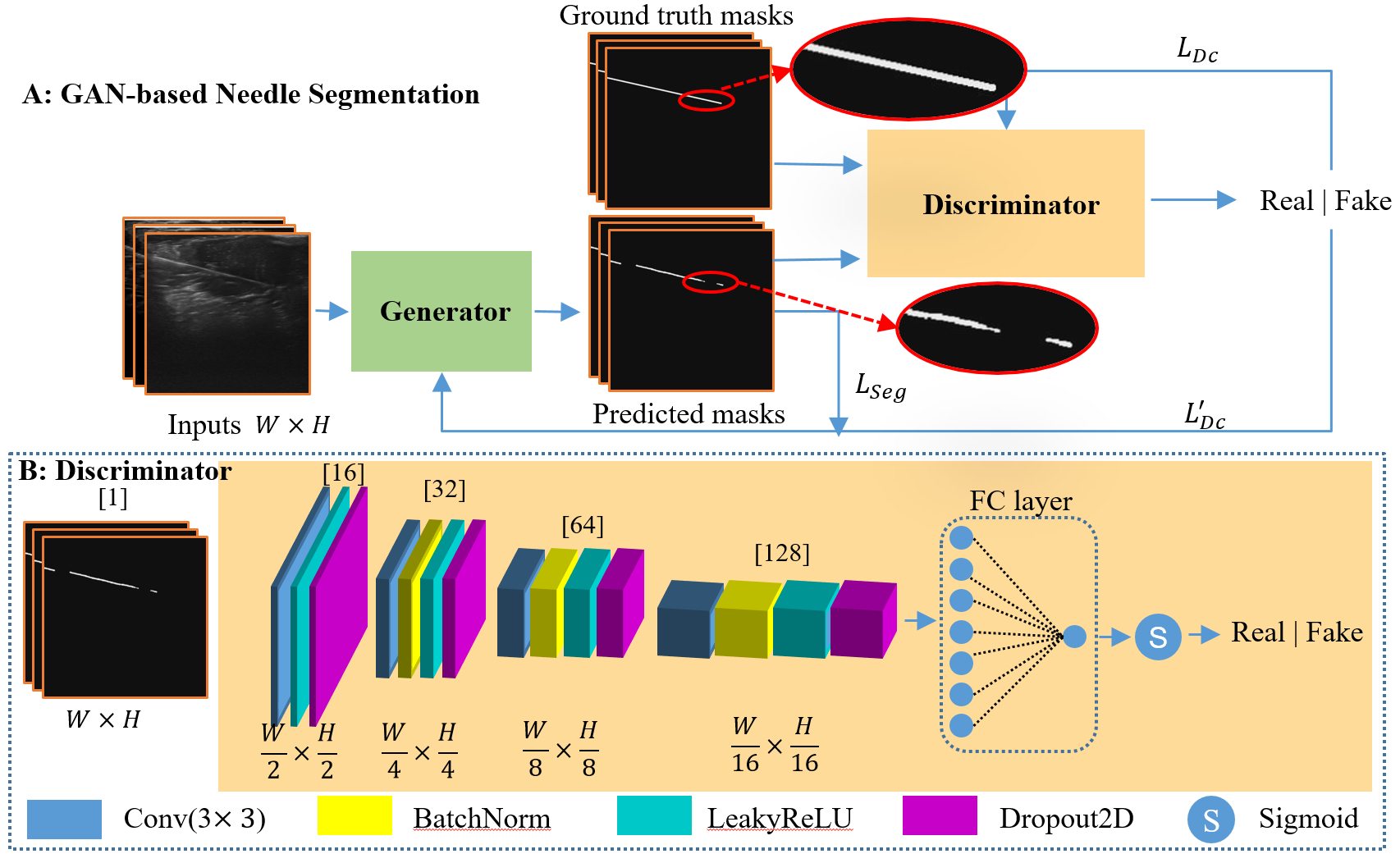}
\caption{(a) Illustration of the proposed AdvSeg-Net for thin instruments segmentation in US images. (b) Detailed structure for the discriminator. }
\label{fig:gen_us_net}%
\end{figure*}

\subsection{Network Architecture}
\par
In this study, an adversarial architecture is employed to train a needle segmentation model for US images. \revision{This is a typical semantic segmentation task where each pixel is assigned to either the binary needle mask or the background.}
The proposed architecture follows the standard GAN layout, comprising a Generator $G$ and a discriminator $D$~\cite{luc2016semantic}. The Generator is a segmentation network using B-mode images as inputs. \revision{By training an adversarial network to discriminate input masks either from the generator predictions or from the ground truth, the networks are compelled to enhance high-order consistencies between the segmentation results and the ground truth in an unsupervised mode. 
In our context, the segmentation network is enforced to enhance the precision of segmenting the imbalanced needle from US images by aligning the segmented mask with ground truth.} The overall structure of the proposed AdvSeg-Net for needle segmentation is depicted in Fig.~\ref{fig:gen_us_net}~(A) and the discriminator is depicted in Fig.~\ref{fig:gen_us_net}~(B), respectively.  

\par
Regarding semantic segmentation tasks, a number of notable models have emerged as dominant players, and they have been proven to be effective in diverse tasks of medical image processing. A representative is U-Net~\cite{ronneberger2015u} and its variants (Attention-UNet~\cite{oktay2022attention} and UNet++~\cite{zhou2018unet++}). They have been widely used in US image segmentation and demonstrated promising performance. In addition, DeepLabV3+~\cite{chen2018encoder} is another successful architecture, which is organized with multiple cascaded branches using Atrous Convolution with diverse fields of perception. In this structure, both global and local precision can be expected. Based on the ablation study in Section~\ref{sec:experimental_segmentation}, UNet++~\cite{zhou2018unet++} is used as the generator in the final model because of the overall performance in terms of various metrics.  

\par
Acknowledging the effectiveness and stability of the training process, as well as the capability to generate realistic images in Deep Convolutional GAN (DCGAN)~\cite{radford2015unsupervised}, the key components of the discriminator in our model are designed following the principles of DCGAN. Stride convolutions are employed instead of pooling operations, and batch normalization is implemented in all layers except the first one. This design is used to reduce the oscillation and improve the stability of the training process of an adversarial network. The activation function is LeakyReLU function. The detailed structure of the used discriminator is depicted in Fig.~\ref{fig:gen_us_net}~(B).

\subsection{Loss Function}~\label{sec:loss}
\par
Due to the fact that the needle is relatively small in comparison with the background, a suitable loss function plays a key role in correctly learning the statistic model. In this study, the imaged area is around $2500~mm^2$ (transducer footprint length: $51.3~mm$, image depth: $50~mm$), while the needle only covers approximately $2-50~mm^2$ (diameter of 18G needle: $1.3~mm$) at different phases during the needle insertion. This leads to a class imbalance of $1:50$. In the deep layers of the generator, the needle appearance will be further weakened due to the downsampled size of the feature map. Inspired by the use of hybrid loss in~\cite{yang2021efficient}, we explore the effectiveness of various loss function combinations in our scenarios. We investigated the performance of cross entropy loss (CE)~\cite{bertels2019optimizing}, focal loss (FL)~\cite{lin2017focal}, and Dice loss~\cite{milletari2016v} to train the generator. Then, the performance of further incorporating the adversarial loss (Adv) is studied by examining the importance of regions in images using Gradient-weighted Class Activation Mapping (Grad-CAM)~\cite{selvaraju2017grad}. 

\par
To properly guide the learning of the proposed AdvSeg-Net, the loss mainly used to train the generator is termed as $\mathcal{L}_{Seg}$. This loss is used to encourage the segmentation model to accurately predict the correct label for each pixel. The second loss is an adversarial loss $\mathcal{L}_{Adv}$ to fine-tune the generator's parameters, aiming to distinguish between the predicted network and ground truth maps. If the discriminator identifies inconsistencies in high-order statistics between the predicted results and ground truth, $\mathcal{L}_{Adv}$ will be used to penalize the generator.

\subsubsection{Segmentation Loss Function}
\par
Regarding medical binary segmentation tasks, CE is one of the most intuitive loss functions to ensure prediction shares the same statistics as ground truth. To assign a relatively higher weight to the regions of interest, FL~\cite{lin2017focal} is employed to train the image segmentation network. Both CE and FL are pixel-wise metrics. Inconsistent or imprecise annotations will pose a significant impact on the regression of the model. Our task of needle segmentation makes it very hard to have an accurate annotation of the thin object's boundary on noisy B-mode images. To balance the uncertainty across all annotations, the region-based Dice loss is used in this study to train the generator. The ablation studies using other losses are also provided in Section~\ref{sec:experimental_segmentation}. 


\begin{equation}\label{eq_dice_loss}
\mathcal{L}_{Dice} = 1 - \frac{2~|\hat{Y}\cap Y|}{|\hat{Y}|+|Y|}
\end{equation}
where $Y$ is the labeled images in which the object is carefully annotated, and $\hat{Y}$ is the predicted mask indicating the target object position in US images.

\par
Unlike other tasks, such as bone segmentation~\cite{alsinan2020bone}, where the boundary is relatively clear in US images, this study faces difficulties due to severe class imbalance between needles and the background. Coupled with the presence of US-specific artifacts and attenuation, obtaining precise annotations becomes challenging. To alleviate the negative impacts caused by imprecise annotation on accurate predictions, we employ the contextual loss~\cite{mechrez2018contextual, yang2021efficient} to encourage the learning of overall contextual information. This loss aims to maintain a robust feature representation by considering not only the individual pixel within the needle but also its surroundings. The definition of the contextual loss $\mathcal{L}_{CL}$ is given as follows:


\begin{equation}\label{eq:contextual_loss}
    \mathcal{L}_{CL}(Y, \hat{Y}) = 1 - \text{cos}\left[\theta_{VGG19}(Y), \theta_{VGG19}(\hat{Y})\right]
\end{equation}
where $Y$ and $\hat{Y}$ are the ground truth and predicted masks. $\theta_{VGG19}(\cdot)$ represents the pre-trained VGG19 model~\cite{simonyan2014very} on ImageNet~\cite{deng2009imagenet}. $\text{cos}(\cdot)$ computes the cosine similarity between two feature maps extracted by the VGG19 encoder. Then, the $\mathcal{L}_{Seg}$ can be defined as follows:

\begin{equation}\label{eq:segmentation_loss}
    \mathcal{L}_{Seg} = \lambda_{Dice}\mathcal{L}_{Dice} + \lambda_{CL} \mathcal{L}_{CL}
\end{equation}
where $\lambda_{Dice}$ and $\lambda_{CL}$ are the weights of $\mathcal{L}_{Dice}$ and $\mathcal{L}_{CL}$. In our scenario, the experimental results demonstrated that a good segmentation performance can be achieved when $\lambda_{Dice}:\lambda_{CL} = 1:0.001$. 

\subsubsection{Adversarial Loss Function}
\par
The dice loss and contextual loss have been employed to optimize the parameters of the segmentation module in supervised mode. In addition, to encourage the segmentation network to learn the consistency and continuity in high-order feature maps, a discriminator is designed to identify the inconsistencies between predictions and ground truth. Therefore, this discriminator can be used to force the generator to create a relatively realistic result as the ground truth in an unsupervised mode. The efficiency of employing \revision{Generative Adversarial Networks (GAN)} to enhance segmentation performance has been demonstrated in the context of \revision{Magnetic Resonance Imaging (MRI)} spine segmentation~\cite{han2018spine} and US shadow segmentation~\cite{alsinan2020bone}. However, its performance in thin instrument segmentation has not been investigated, particularly considering the inevitable noise and speckle in US images.

\par
Following the theory of two-player minimax game, the objective of the generator network is to minimize the probability of the generated masks being recognized while maximizing the probability of the discriminator making a mistake. The discriminator loss $\mathcal{L}_{Adv}$ is defined as follows:

\begin{equation} \label{eq:loss_DC}
   \mathcal{L}_{Adv} =  \frac{1}{N}\sum_{i-1}^{N}
   \mathcal{L}_{bce}\left[D(G(X_i)),0\right] + \mathcal{L}_{bce}\left[D(Y_i),1\right]
\end{equation}
where $X$ is the input B-mode images, $G(\cdot)$ is the predicted mask using the generator, $Y$ represents the ground truth.  

\par
To trick the discriminator, the segmentation masks should be close to the ground truth to avoid being distinguished. To regulate the parameter tuning of the generator, we only need to consider whether the discriminator can successfully identify fake images. The loss for generator after combining adversarial loss $\mathcal{L}_{Gen}$ is defined as follows:

\begin{equation} \label{eq:loss_Gen} 
\begin{split}
    \mathcal{L}_{Gen} &= \mathcal{L}_{Seg} + \lambda_{Adv} \mathcal{L}_{Dc}^{\prime} \\
                    &= \mathcal{L}_{Seg}+\lambda_{Adv}\frac{1}{N}\sum_{i-1}^{N} \mathcal{L}_{bce}\left[D(G_{\theta}(X_i)),1\right]
\end{split}
\end{equation}
where $\lambda_{Adv}$ is the weight for adversarial loss. In this study, $\lambda_{Adv}$ is $0.1$ based on the experimental performance.

\subsection{Training}
\subsubsection{Data Augmentation}
\par
\revision{Data Augmentation has become a standard solution to improve the performance of CNNs and expand limited datasets to take advantage of the capabilities of big data~\cite{shorten2019survey}.} 
In this study, data augmentation is applied during the training to enhance robustness and reduce potential overfitting~\cite{shorten2019survey}. The detailed augmentation parameters are listed in Table~\ref{tab:augmentations_needle}. Throughout the training process, the specified augmentation methods are randomly combined sequentially, guided by preset probabilities. The horizontal flipping can enhance the model's robustness to variations in needle insertion directions. To consider brightness variations caused by varying settings or diverse machines, the image brightness contrast is randomly adjusted in the range of $[-10\%,10\%]$ with a probability of 30\%. Furthermore, by adding Gaussian noise and incorporating blurring operations, the trained model could be more resilient to random speckles and noise in 2D US images.

	\begin{table}[htb!]
		\centering
		\caption{Applied Image Augmentation Parameters}%
		\label{tab:augmentations_needle}
		\begin{tabular}{lc}%
		\noalign{\hrule height 1.2 pt}
			 \bfseries{Augmentation Method}    & \bfseries{Probability} \\
			 \hline
			 Horizontal Flip  & 30\% \\
			 Contrast ($\pm10\%$ ) & 30\% \\ 
			 Gaussian Noise  & 20\% \\
			 Gaussian Blurring  & 20\% \\
		\noalign{\hrule height 1.2 pt}
		\end{tabular}
	\end{table}

\subsubsection{Training Details}
\par
The network was trained on a mobile workstation (GPU: GeForce RTX 4080 Mobile, CPU: Intel i9-12900HX). AdvSeg-Net is structured in an adversarial fashion. It is challenging to find the delicate balance between optimizing the generator, which aims to create realistic segmentation masks, and the discriminator, which endeavors to differentiate between masks generated by the generator and ground truth. Since the discriminator is employed to encourage the generator to create the masks that preserve consistency and continuity in high-order feature maps, we conduct the training of AdvSeg-Net in two separate phases. To have a stable and good parameter initialization from scratch, we initially train the generator purely based on $\mathcal{L}_{Seg}$ [Eq.~(\ref{eq:segmentation_loss})]. In this stage, the training details, including the number of training epochs, batch size, and learning rate, are set to $40$, $8$, and $1\times10^{-4}$, respectively.

\par
After achieving a good initialization of the generators, the second phase involves the joint update of the parameters for both the generator and discriminator. This is accomplished using $\mathcal{L}_{Gen}$ [Eq.~(\ref{eq:loss_Gen})] and $\mathcal{L}_{Adv}$ [Eq.~(\ref{eq:loss_DC})] respectively. The parameters of the generator and discriminator are updated alternately. To avoid excessively large fluctuations caused by unconverged discriminator, the learning rate of the discriminator ($1\times10^{-5}$) is set to be relatively smaller than that for the generator ($5\times10^{-5}$). Additionally, both learning rates decrease by half every $8$ epochs. The other training details include the training epoch of $40$, and a batch size of $8$. The optimization of parameters is carried out using Adam~\cite{kingma2014adam} as the default optimizer throughout the study.

\par
To identify proper hyperparameters, the open-source tool Optuna\footnote{https://opencv.org/} was employed to determine the importance of six hyperparameters: weights of $\lambda_{Dice}\in[0.5,~1]$, $\lambda_{CL}\in[0.0001,~0.1]$ and $\lambda_{Adv}\in[0.01,~0.5]$, learning rates of generator $l_{gen}\in[10^{-5}, 10^{-2}]$ and discriminator $l_{Adv}\in[10^{-5}, 3^{-3}]$. The batch size is also considered, with options: $\{4, 8\}$. To find the optimal configuration, $200$ trials were autonomously conducted, exploring hyperparameter combinations randomly selected within the specified ranges. The results emphasize the significant importance of $l_{Adv}$, with a weight of 0.427, surpassing the relevance of other hyperparameters. This observation aligns with well-documented challenges in training GANs, where a large $l_{Adv}$ during the early training stages can lead to instability and fluctuations in the training process.

\section{Needle Determination and Robotic US Probe Re-positioning}

\begin{figure*}[ht!]
\centering
\includegraphics[width=0.95\textwidth]{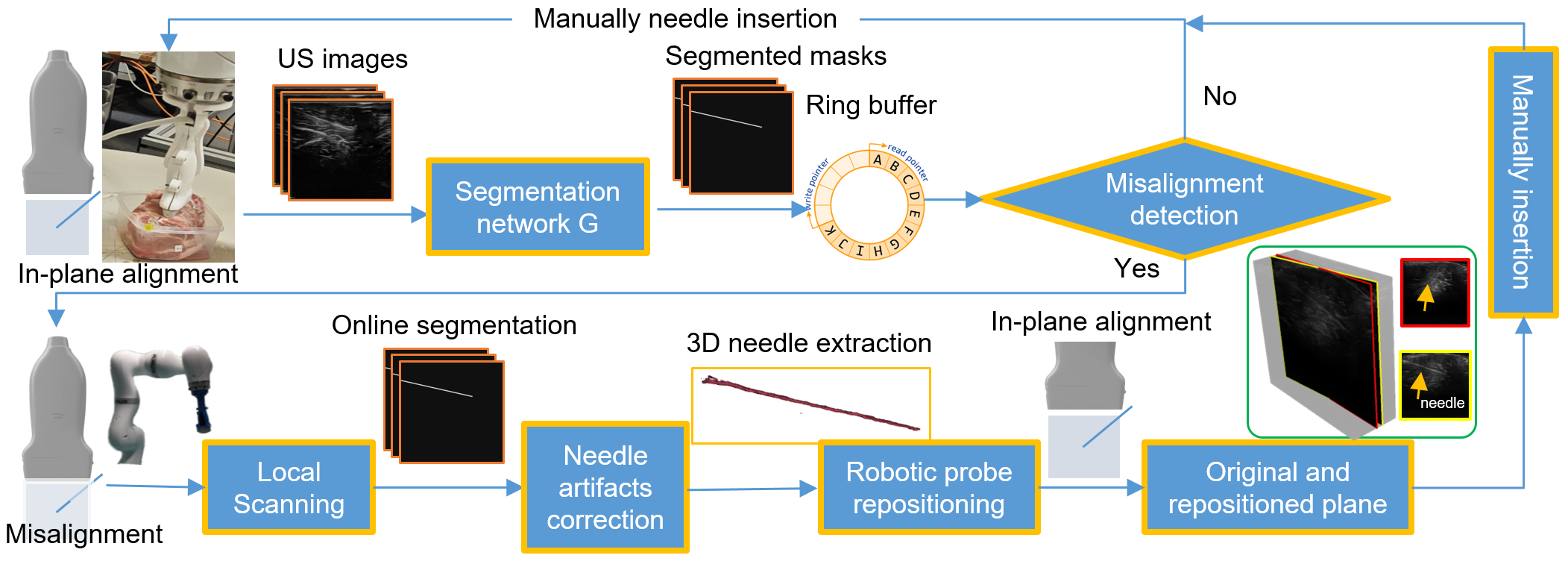}
\caption{The schematic illustration of the probe repositioning. }
\label{fig:robot_repositon_overview}%
\end{figure*}

\par
This section presents an autonomous re-localization method aiming to maintain the visibility of the needle in US images. To do so, a segmentation-based misalignment monitoring module is designed. Once misalignment occurs, a local search in transverse scanning mode will be triggered to re-localize the inserted needle. Real-time segmentation will be computed using the trained network in Section~\ref{sec:network_architecture}, and the slice thickness artifacts will be mitigated to generate a line-like needle shaft in 3D. Based on the results of the stacked needle masks with tracking information, the translational and angular movements needed to restore the needle visibility in 2D images can be determined. The schematic of the robotic probe repositioning pipeline has been depicted in Fig.~\ref{fig:robot_repositon_overview}. A representative local 3D volume is depicted in gray in Fig.~\ref{fig:robot_repositon_overview} with two US planes representing the original plane (red) and the one after repositioning (yellow).

\subsection{Misalignment Detection During Needle Insertion}~\label{sec:inplane_detection}
\subsubsection{Needle Determination from Segmentation Masks}
The insertion mode refers to the needle being well aligned within the in-plane view. In this mode, the whole inserted needle, including its tip, can be completely visualized in each frame. To extract the needle, the trained segmentation network (the generator in Fig.~\ref{fig:gen_us_net}) is initially employed to perform the online segmentation for real-time B-mode images. Then, a set of post-processing procedures are applied to determine both the needle shaft and tip based on the segmented masks. Considering the potential of non-perfect segmentation: (1) we first compute the closing contour of each disconnected area on the segmented mask and compute the minimum enclosing rectangle $S_{Rec}$ of the extracted contours; (2) Then, the largest $S_{Rec}$ is considered as the target or the main component of the target $S_{tag}$. Its centroid is marked as $C_{tag}\in R^{2\times 1}$; (3) To remove the potential outlier, we calculate the slope angle $\theta_{slo}^{i}$ of the line connecting the points $C_{tag}$ and $C_{i}$ of the $i$-th enclosing rectangle $S_{Rec}^{i}$. If the computed $\theta_{slo}^{i}$ deviates from $\theta_{slo}^{tag}$ (fitted line of $S_{Rec}$) over
a given threshold $\theta_{slo}^{th}$, $S_{Rec}^{i}$ is considered as outlier. Otherwise, $S_{tag}$ will be updated as $S_{tag}=  \{S_{tag}, S_{Rec}^{i}\}$. By repeating the aforementioned processes, we can obtain the final $S_{tag}$ representing the complete needle shaft. (4) finally, a line-fitting algorithm is implemented to fit the segmented pixels included in the final $S_{tag}$. The tip is represented by the endpoint of the fitted line.

\subsubsection{Misalignment Monitoring}~\label{sec:misalignment}
Misalignment is undesirable as it can result in operators losing track of the inserted needle. Close monitoring of inserted needles is crucial for clinical applications to avoid the risk of damaging the vital adjacent organs, e.g., kidney or liver drainage procedures. To detect misalignment between the probe and the inserted needle, the length of the determined shaft in the last subsection is stored in a ring buffer. Since the human tissues are soft, the needle shaft will disappear from US images when the inserted needle deviates from the US plane. Therefore, if the current length deviates from the average length of the last $N_{ring}$ frames in the ring buffer by more than $T_{mis}$, it is indicative of misalignment. To balance the sensitivity and stability in this detection, $N_{ring}$ is set to $25$, and $T_{mis}$ is set to $40\%$ based on experimental performance.

\subsection{Slice Thickness Correction and Needle Determination in Transverse Scanning}~\label{sec:localization}
\par
\begin{figure*}[ht!]
\centering
\includegraphics[width=0.85\textwidth]{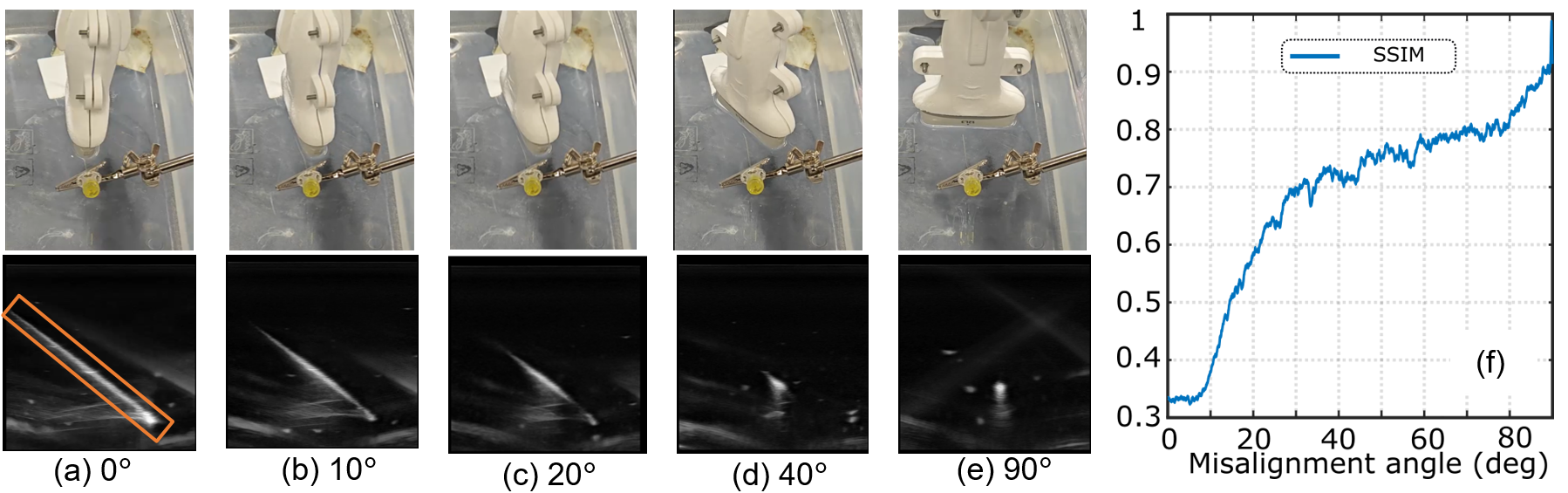}
\caption{Illustration of US slice thickness artifact with respect to varying $\theta_{mis}$. This representative result is obtained using a water tank. (a)-(e) The real setup and corresponding US images when $\theta_{mis}$ is $0$, $10^{\circ}$, $20^{\circ}$, $40^{\circ}$, and $90^{\circ}$, respectively. (f) The computed SSIM between the last frame and all other frames. The orange rectangle is a pre-defined region of interest to limit the negative impact caused by irrelevant US artifacts.}
\label{fig:artifact_correction}%
\end{figure*}

During transverse scanning, the needle is positioned out-of-plane relative to the probe. Ideally, a bright point at the intersection point between the needle and the in-plane slices should be observed. However, it is worth noting that the effect of US slice thickness artifacts will result in the visualization of a ``shot line" rather than a single "intersection point" when the needle is out-of-plane (see Fig.~\ref{fig:artifact_correction}). \revision{It is also noteworthy that similar phenomena can be observed in ex vivo animal tissues.}

\par
This image artifact is commonly referred to as ``slice thickness"~\cite{goldstein1981slice}, arising from the non-ideal 2D US image plane, which has a certain thickness. Within this assumed ``thick" scan plane (elevation beam width), all echoes are mistakenly interpreted as originating from structures within the presumed ``thin" scan plane. In other words, the final 2D US images are generated by projecting the short segment of the needle in the small 3D volume, formed by the elevation beam width and US view plane, onto the 2D US image plane. Such an artifact leads to a significant impact on tasks related to needle positioning and visualization~\cite{peikari2012characterization, peikari2011effects}. It can be seen from Fig.~\ref{fig:artifact_correction} that a ``shot line" rather than a single "intersection point" will appear when the probe deviates from the ideal in-plane alignment. In this experiment, we rotated the probe along its centerline. The rotated angle count from in-plane alignment is termed misalignment angle $\theta_{mis}$. It can be seen from Fig.~\ref{fig:artifact_correction} that the needle appearance in US images approaches that of a point as $\theta_{mis}$ increases. To quantitatively trace this change, we computed the Structural Similarity Index (SSIM) between the last frame ($\theta_{mis}=90^{\circ}$) and individual images obtained when $\theta_{min}\in[0,90^{\circ}]$ in Fig.~\ref{fig:artifact_correction}~(f). The results demonstrated that the US needle appearance remains similar to the ideal case when $\theta_{mis}$ is smaller than $7^{\circ}$. Subsequently, its appearance linearly changes between $8^{\circ}$ to $40^{\circ}$. Beyond $40^{\circ}$, the needle appearance closely resembles a point [see Fig.~\ref{fig:artifact_correction}~(d)] and gradually approaches the intersection point as $\theta_{mis}$ increases.

\par
While this artifact alters the appearance of the needle, the magnified needle in US images (from a point to a short line) can aid in locating the needle for 2D segmentation. Since we lack access to the beamforming shape parameter of commercial US machines, we employ the center point of the segmented short line to represent the interaction point in all 2D slices generated during the transverse local scan. 
Building upon the spatial and temporal US-robot calibration process detailed in~\cite{jiang2021autonomous, jiang2023robotic}, we are able to map each pixel in US images to a fixed 3D world coordinate system. This involves stacking the segmented intersection points in each slice based on corresponding robotic tracking information in 3D space. \revision{To enhance the accuracy, the explicit outliers can be removed using Density-Based Spatial Clustering of Applications with Noise (DBSCAN)~\cite{ester1996density} by excluding points with few neighbors within a specified sphere around them.}
To further generate a continuous line from the segmented sparse needle intersection points in individual slices, the Line-RANSAC algorithm~\cite{Mariga_pyRANSAC-3D_2022} is applied to compute the fitted line parameters. These parameters include the coordinates of two points that determine the linear axis of the needle.

%

\subsection{US Probe Repositioning}~\label{sec:Repositioning}

\begin{figure}[htb!]
\centering
\includegraphics[width=0.45\textwidth]{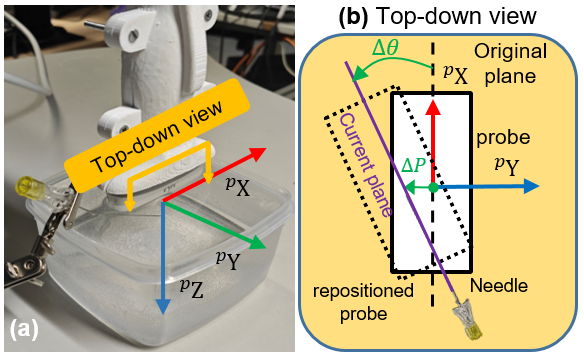}
\caption{(a) Illustration of the real scene. (b) Top-down view along the Z axis of probe frame $\{p\}$. The solid and dashed rectangles represented the original and adjusted probe pose, respectively. 
}
\label{Fig_probe_reposition}
\end{figure}

\par
To properly visualize the tip of the misaligned needle, adjustments are required for both translational and rotational displacements, denoted as $\Delta \textbf{P}$ and $\Delta \Theta$, respectively. It is worth noting that the translational adjustment is performed along the probe short axis ($^{p}Y$ of probe frame $\{p\}$ in Fig.~\ref{Fig_probe_reposition}) rather than in the direction orthogonal to the current needle plane. This design is used to mitigate potential interference or collisions between the probe and the portion of the needle extending beyond the contact skin. Based on the extracted 3D needle (described in robotic base frame $\{b\}$) from Section~\ref{sec:localization}, we randomly selected two points ($^{b}P_1$, and $^{b}P_2$) located on the extracted needle shaft. Since the local adjustment is more intuitively in the probe frame $\{p\}$ as shown in Fig.~\ref{Fig_probe_reposition}, the two points are projected to the $X\text{-}Y$ plane of frame $\{p\}$ (${^{p}P_1^{\prime}}= {^{p}_{b}T} ^{b}P_1$, and ${^{p}P_2^{\prime}}= {^{p}_{b}T} ^{b}P_2$). Then, $\Delta \textbf{P}$ and $\Delta \Theta$ are computed as follows:

\begin{equation} ~\label{eq:reposition}
\begin{split}
    &\Delta\textbf{P} = \left[0, {^{p}P_2^{\prime}}(1)-{^{p}P_2^{\prime}}(0)\frac{{^{p}P_1^{\prime}}(1)-{^{p}P_2^{\prime}}(1)}{{^{p}P_1^{\prime}}(0)-{^{p}P_2^{\prime}}(0)}, 0 \right]^{T} \\
    &\Delta \Theta =\left[0, 0, \text{arctan2}\left(\frac{\|{^{p}\textbf{X}} \times \overrightarrow{{^{p}P_1^{\prime}~ {^{p}P_2^{\prime}}}}\|}{ {^{p}\textbf{X}} \cdot \overrightarrow{{^{p}P_1^{\prime}~ {^{p}P_2^{\prime}}}}}\right) \right]
\end{split}
\end{equation}
where $P(i)$, $i \in \{0,~1,~2\}$ represents value in X, Y, and Z dimensions, respectively. $\Theta(i)$ is the rotation along X, Y, and Z axis, respectively. \text{arctan2} computes the signed angle whose tangent is the quotient of its arguments. The angle is counted from $^{p}\textbf{X}$ to $\overrightarrow{{^{p}P_1^{\prime}~ {^{p}P_2^{\prime}}}}$. To ensure the consistence of the signs, ${^{p}P_1^{\prime}}$ need to be smaller than ${^{p}P_2^{\prime}}$ in $^{p}\textbf{X}$ direction. The computed $\Delta \textbf{P}$ and $\Delta \Theta$ can be directly applied to the current probe pose (solid rectangle in Fig.~\ref{Fig_probe_reposition}) to the new probe pose (dashed rectangle in Fig.~\ref{Fig_probe_reposition}).

\section{Results}
\subsection{Experimental Setup and Dataset}
\par
The experimental setup mainly comprises a robotic manipulator (LBR iiwa 14 R820, KUKA GmbH, Germany) and an ACUSON Juniper US machine (Siemens Healthineers, Germany). A liner US probe (12L3, Siemens Healthineers, Germany), with a footprint length of $51.3~mm$, was attached to the robotic flange. To manipulate the robotic system, a self-developed Robot Operating System (ROS) software~\cite{hennersperger2016towards} was employed. To access US images, we utilized a frame grabber (Epiphan Video, Canada) via OpenCV. The frequencies for robot control and image acquisition are $100~Hz$ and $30~fps$, respectively. Without further specification, the default settings for US acquisition are as follows: image depth was $50~mm$ with a focus depth of $30~mm$; Tissue Harmonic Imaging (THI) was $6.7~MHz$; Dynamic Range (DynR) was $80~dB$. 

\par
To mimic the US image appearance of human tissue, four distinct samples of ex vivo porcine tissue were used for data recording. Comprising with bovine tissue and gelatin phantom, porcine tissues can introduce more noisy artifacts due to the fat, muscle, and rich fascial tissue. The images were recorded in both insertion mode and transverse scanning mode. It is noteworthy that the images obtained in the transverse scanning mode are also inclined to be a line rather than a single point in US images due to the slice thickness artifact. So, only a single AdvSeg-Net model is trained. In total, we recorded $12$ sweeps on different porcine tissues. To reduce the similar images in neighboring frames, all sweep was downsampled. For training, each sweep contains $180-300$ images. \revision{The dataset consists of a total of 2535 images: 1761 images from 8 sweeps in insertion mode and 774 images from 4 sweeps in transverse mode. The ratio between training and validation is 8:2.} In addition, for testing, we recorded $8$ sweeps in the insertion mode and $5$ sweeps in the transverse scanning mode on an unseen ex vivo porcine sample. \revision{The image size is $671\times 657$ pixels, and all the images are carefully labeled by US experts.}

\subsection{Needle Segmentation Performance}~\label{sec:experimental_segmentation}
\par
To evaluate the segmentation performance of the needle, we compute both image-wise segmentation metrics (Recall, Precision, IOU, and continuity), and application-wise needle detection metrics (tip error and angle error). 
The tip position error $e_{tip}$ and angle error $e_{\theta}$ in the insertion mode are computed between the carefully annotated label and the detected needle in individual slices. The center error is computed based on the distances from the image center point to the predicted needle shaft and annotated needle shaft~\cite{lee2020ultrasound}.

\par
Recall, Precision, and IOU scores are standard metrics for assessing the segmentation performance. Specific to our task of needle segmentation, the procedures for computing the continuity of the needle segmentation results are summarized as follows. First, to restrain negative impacts caused by the potential outliers in segmentation results, the ground truth masks are used to determine the minimum enclosing rectangle as the ROI. Then, we assign a binary value along the needle axis. Iterating through each short line in parallel to the short axis of the ROI, the false value ($0$) will be set if the ratio between the total number of pixels belonging to the predicted mask and the length of the short axis of ROI is smaller than a threshold $T_{con}$. Otherwise, the true value ($1$) will be set. \revision{Finally, the continuity is computed as $\frac{N\{P| p = 1\}}{N_{long}}$, where the numerator is the number of locations with value one and the denominator is the number of pixel of the ROI's long axis. In this study, $T_{con}$ is $70\%$.} The results computed on $400$ images from eight new US sweeps ($50$ per sweep) obtained on an unseen sample of porcine tissue are summarized in Table~\ref{tab:segmentation_needle_insertaion}.


\begin{table*}[t]
\begin{center}
\caption{Segmentation Performance Using Different Approaches for Insertion Mode (mean $\pm$ std)}
\label{tab:segmentation_needle_insertaion}
\resizebox{0.96\textwidth}{!}{
\begin{tabular}{cccccccccccccc}
\noalign{\hrule height 1.2 pt}
\multirow{2}{*}{Network}  &  \multicolumn{4}{c}{Image-wise segmentation metrics} & & \multicolumn{3}{c}{Application-wise needle detection metrics} & \multirow{2}{*}{Time (ms)} \\
\cline { 2 - 5 } \cline{7-9} 
& Recall & Precision & IOU  & Continuity & & Tip error (mm) & Angle error ($^{\circ}$) & Center error (mm) & \\
\hline
U-Net w CE          & $0.74\pm0.11$ & $0.69\pm0.13$ & $0.56\pm0.13$ & $0.88\pm0.12$ && $0.34\pm0.29$ & $2.98\pm3.88$ & $0.17\pm0.35$ & 4.3 (1.3)\\
U-Net w FL          & $0.75\pm0.10$ & $0.70\pm0.12$ & $0.57\pm0.12$ & $0.90\pm0.10$ && $0.33\pm0.27$ & $2.55\pm3.48$ & $0.13\pm0.11$ & 4.3 (1.3)\\
U-Net w Dice        & $0.79\pm0.08$ & $0.77\pm0.10$ & $0.64\pm0.11$ & $0.92\pm0.08$ && $0.32\pm0.26$ & $1.75\pm2.30$ & $0.12\pm0.10$ & 4.3 (1.3)\\
U-Net w FL+CL       & $0.77\pm0.07$ & $0.78\pm0.15$ & $0.63\pm0.13$ & $0.89\pm0.14$ && $0.35\pm0.79$ & $3.11\pm4.96$ & $0.11\pm0.17$ & 4.3 (1.3)\\
U-Net w Dice+CL     & $0.80\pm0.08$ & $0.75\pm0.10$ & $0.64\pm0.11$ & $0.92\pm0.09$ && $0.30\pm0.24$ & $1.88\pm2.74$ & $0.11\pm0.10$ & 4.3 (1.3)\\
\hline
Attention-UNet      & $0.76\pm0.06$ & $0.68\pm0.11$ & $0.56\pm0.09$ & $0.88\pm0.10$ && $0.51\pm0.53$ & $2.04\pm2.06$ & $0.12\pm0.17$ & 4.9 (1.8) \\
UNet++              & $0.77\pm0.05$ & $0.78\pm0.06$ & $0.63\pm0.06$ & $0.96\pm0.04$ && $0.37\pm0.29$ & $1.33\pm0.92$ & $0.10\pm0.09$ & 6.4 (2.6) \\
DeeplabV3+          & $0.71\pm0.04$ & $0.77\pm0.07$ & $0.58\pm0.06$ & $0.98\pm0.03$ && $0.38\pm0.33$ & $1.02\pm0.84$ & $0.11\pm0.13$ & 5.0 (1.9) \\
AdvSeg-Net$^{(1)}$  & $0.74\pm0.05$ & $0.77\pm0.05$ & $0.61\pm0.06$ & $0.98\pm0.02$ && $0.43\pm0.37$ & $0.98\pm0.87$ & $0.13\pm0.15$ & 5.0 (1.9) \\
AdvSeg-Net$^{(2)}$  & $0.76\pm0.04$ & $0.79\pm0.05$ & $0.63\pm0.05$ & $0.97\pm0.03$ && $0.37\pm0.29$ & $1.19\pm0.85$ & $0.10\pm0.10$ & \revision{6.4 (2.6)} \\
\hline
\noalign{\hrule height 1.2 pt}
\end{tabular}
}
\end{center}
{\textit{*Nomenclature:}
AdvSeg-Net$^{(1)}$ and AdvSeg-Net$^{(2)}$ employ DeeplabV3+ and UNet++ as the generator, respectively.
}
\end{table*}

\subsubsection{Effectiveness of Loss Functions}
\par
Segmenting thin objects, such as needles, from 2D US images poses a challenge due to the presence of biological boundaries within in vivo tissues. These boundaries exhibit strong contrast due to the reflection of US signals at such boundaries, creating a visual resemblance to thin objects in US images. Before investigating the performance of different network architectures, we first assess the impact of various loss functions on performance variation. To provide a fair comparison, the standard U-Net is used as the backbone. Then, the typical loss functions CE, FL, and Dice are investigated. Considering the class imbalance, we also compared the hybrid loss combining CL to FL and Dice, respectively. The results on five unseen insertion sweeps obtained on unseen ex vivo porcine samples are summarized in Table~\ref{tab:segmentation_needle_insertaion}.

\par
\begin{figure}[htb!]
\centering
\includegraphics[width=0.48\textwidth]{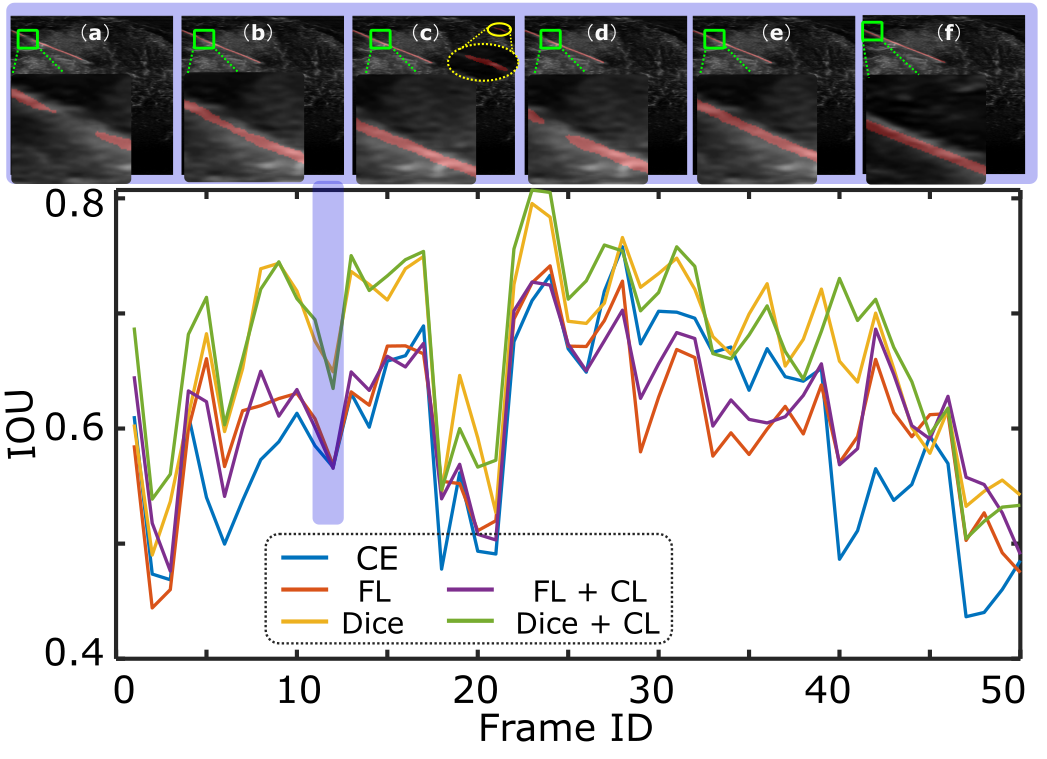}
\caption{Illustration of the computed IOU value for an unseen sweep obtained on an unseen porcine ex vivo sample. (a)-(e) are the segmentation results of frame $12$ using CE loss, FL, Dice, FL+CL and Dice+CL, respectively. (f) is the manual annotation of the same frame.  
}
\label{Fig_iou_figure}
\end{figure}

From the first two rows in Table~\ref{tab:segmentation_needle_insertaion}, FL can result in a slightly better performance than the standard CE in both image-wise segmentation metrics and application-wise metrics. However, it is noteworthy that the improvement is small (around $1\%$ for all image-wise metrics). Compared to CE and FL, relatively large enhancements are witnessed using Dice loss (recall: $0.74$, $0.75$, and $0.79$; precision: $0.69$, $0.70$, and $0.77$; and IOU: $0.56$, $0.57$, and $0.64$). This significant performance enhancement is because Dice is computed based on the whole object, whereas CE and FL are calculated pixel by pixel. Due to the inevitable inaccuracy of labeling such a thin object [see Fig.~\ref{Fig_iou_figure}~(f)], CE and its variant FL will be confused by the pixels being wrongly annotated. 

\par
Since the target class and background class are severely imbalanced, the understanding of the context besides the thin needle itself is also important for enhancing the overall performance. So VGG-19-based CL loss in latent space is computed as Eq.~(\ref{eq:contextual_loss}). The performance of two hybrid loss \{FL, CL\} and \{Dice, CL\} are summarized in the fourth and fifth rows in Table~\ref{tab:segmentation_needle_insertaion}. After adding CL to FL, significant improvements are demonstrated across recall ($0.75$ vs $0.77$), precision ($0.70$ vs $0.78$), and IOU ($0.57$ vs $0.63$). The improved results are comparable to the results obtained by using Dice. After adding CL to Dice, the slight improvements lead to achieving the best performance in terms of most metrics (both image-wise and application-wise metrics). In order to demonstrate the performance variation intuitively, the IOU of an unseen sweep obtained on an unseen ex vivo porcine tissue is depicted in Fig.~\ref{Fig_iou_figure}. It can be seen from Fig.~\ref{Fig_iou_figure} that the Dice and the hybrid loss \{Dice, CL\} outperform their peers, while the difference between them is relatively small. This finding is consistent with the average results shown in Table~\ref{tab:segmentation_needle_insertaion}. Since the annotation of the thin needle in US images is not perfect [see Fig.~\ref{Fig_iou_figure}~(f)], we further check the predicted masks. Fig.~\ref{Fig_iou_figure}~(a)-(e) are the segmentation results of frame $12$ using CE loss, FL, Dice, \{FL, CL\}, and \{Dice, CL\}, respectively. The zoomed figures show that the predicted masks computed using Dice and hybrid loss \{Dice, CL\} align best with the features (high contrast) shown in B-mode images. However, it is noticeable that Dice incorrectly predicted a biological boundary as the needle on the top right part [see Fig.~\ref{Fig_iou_figure}~(c)]. This ablation study demonstrates that the hybrid loss \{Dice, CL\} can yield the best performance, which will be crucial in the next phase for repositioning the probe to properly visualize the needle after misalignment.

\begin{table*}[tb]
\centering
\caption{Overall Performance of the Needle Repositioning}
\label{tab:performance_reposition}
\resizebox{0.75\textwidth}{!}
{
\begin{tabular}{l|cccccccc}
\noalign{\hrule height 1.2 pt}
\multirow{2}{*}{\diagbox{$\Delta \textbf{P}$}{$\Delta \Theta$}} 
& \multicolumn{2}{c}{ $5^{\circ}$} && \multicolumn{2}{c}{ $10^{\circ}$} && \multicolumn{2}{c}{ $15^{\circ}$} \\
\cline { 2 - 3 } \cline { 5-6 } \cline { 8-9 } 
&  $e_{p}$ (mm) &  $e_{\theta}$ ($^{\circ}$) &&  $e_{p}$ (mm) &  $e_{\theta}$ ($^{\circ}$) && $e_{p}$ (mm) &  $e_{\theta}$ ($^{\circ}$) \\
\hline
0 mm & $1.03\pm0.24$ &  $0.76\pm0.68$ &&  $0.98\pm0.30$ &  $0.82\pm0.78$ && $1.18\pm0.19$ &  $1.20\pm0.88$ \\
3 mm & $0.81\pm0.19$ &  $1.40\pm0.74$ &&  $1.06\pm0.11$ &  $1.48\pm0.79$ && $1.75\pm0.56$ &  $1.78\pm0.68$ \\
6 mm & $1.12\pm0.17$ &  $0.93\pm0.62$ &&  $1.81\pm0.39$ &  $1.23\pm0.96$ && $3.88\pm0.52$ &  $1.69\pm0.81$ \\
\noalign{\hrule height 1.2 pt}
\end{tabular}
}
\end{table*}

\subsubsection{Effectiveness of Network Structures on Needle Segmentation}
\par
To further provide a quantitative comparison, the performance of AdvSeg-Net is compared with popular existing segmentation networks, including U-Net~\cite{ronneberger2015u}, Attention-UNet~\cite{oktay2022attention}, DeepLabV3+~\cite{chen2018encoder}, and UNet++~\cite{zhou2018unet++}. To provide a fair comparison, the same hybrid loss \{Dice, CL\} is used to train the existing approaches. Additionally, to validate the efficacy of the second-stage training, which incorporates adversarial loss, a comparison between the outputs achieved using the generator model trained in the first and second stages is performed as well. The results are summarized in Table~\ref{tab:segmentation_needle_insertaion}. 

\begin{figure}[hbt!]
\centering
\includegraphics[width=0.40\textwidth]{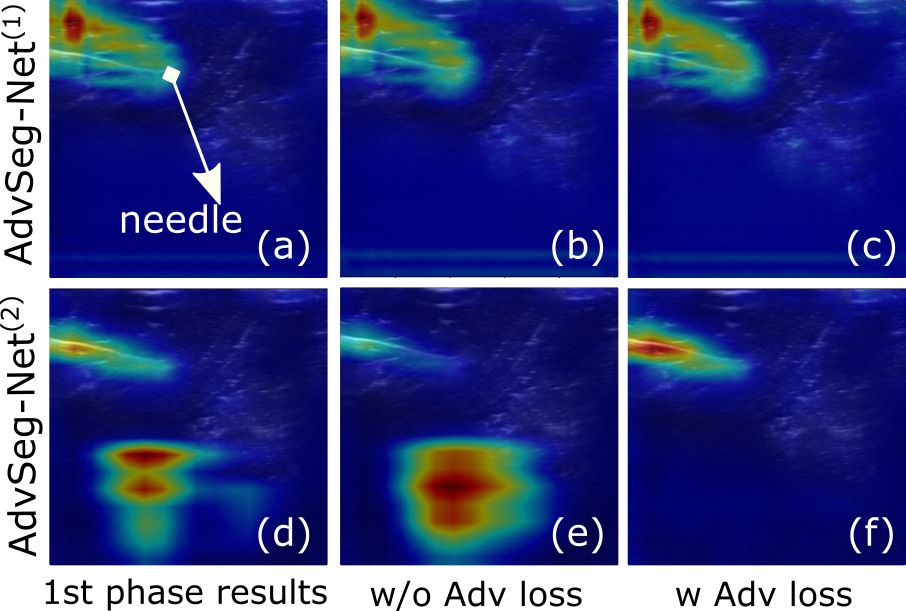}
\caption{Illustration of the important regions for needle prediction using Grad-CAM. (a), (b) and (c) are grad-CAMs for the same image using the DeeplabV3+ trained in the first phase, fine-tuned without and with adversarial loss, respectively. (d), (e) and (f) are the grad-CAMs computed using the UNet++ trained in the first phase, fine-tuned without and with adversarial loss, respectively. 
}
\label{Fig_heatmap}
\end{figure}

\par
Based on the results, we can find that attention-UNet performs worse in comparison to other models. Compared with the standard U-Net, recall, precision, IOU, and continuity are reduced from $0.80$ to $0.76$, from $0.75$ to $0.68$, from $0.64$ to $0.56$, and from $0.92$ to $0.88$, respectively. Similar decrease tendencies are also observed in the application-wise metrics. The results obtained by UNet++ and DeeplabV3+ can outperform the ones obtained using the standard U-Net in some metrics, particularly in continuity. This is because the UNet++ redesigned the skip connection to use deep supervision to ensure the geometrical similarity in the encoder and decoder, while DeeplabV3+ used Atrous Separable Convolution to increase the field of perception. To further validate the effectiveness of the adversarial structure, we use DeeplabV3+ and UNet++ as the generators to form the AdvSeg-Net$^{1}$ and AdvSeg-Net$^{2}$. After obtaining well-trained generators (UNet++ and DeeplabV3+) using only hybrid loss \{Dice, CL\}, the adversarial loss is further incorporated to fine-tune the parameters of the according generators. It can be seen from Fig.~\ref{tab:segmentation_needle_insertaion}, AdvSeg-Net$^{1}$ and AdvSeg-Net$^{2}$ can maintain or slightly increase the performance in terms of metrics. But, it is worth noting that the improvements are not significant. 

\par
To investigate the impact of adversarial loss, we further train the segmentation network that had been trained in the first stage using loss \{Dice, CL\} and loss \{FL, CL\}, respectively. To monitor their effectiveness on needle segmentation tasks, Gradient-weighted Class Activation Mapping (Grad-CAM)~\cite{selvaraju2017grad} is used to visualize the important regions in the image for predicting the objects. A set of Grad-CAM computed on a randomly selected image has been shown in Fig.~\ref{Fig_heatmap}. Fig.~\ref{Fig_heatmap}~{(a)} and (d) are the results obtained using the well-trained DeeplabV3+ and UNet++ segmentation networks in the first phase, while the second and third columns are the ones computed using the fine-tuned models using loss \{Dice, CL\} and loss \{Dice, CL, Adv\}, respectively. Both models were further trained $400$ iterations in the second phase. It can be seen from Figs.~\ref{Fig_heatmap} (a-c) that the marked most important regions are slightly deviating from the real needle objects. For AdvSeg-Net$^{2}$, we can find that it wrongly predicts the regions of importance using the basic UNet++ trained in the first phase [Fig.~\ref{Fig_heatmap}~(d)]. After fine-tuning, the model without adversarial loss still fails to predict the right importance map [Fig.~\ref{Fig_heatmap}~(e)], while the ones obtained by the model fine-tuned with the adversarial model can correctly predict the right needle region [Fig.~\ref{Fig_heatmap}~(f)]. 
Therefore, based on these experimental results, we conclude that AdvSeg-Net$^{2}$ with UNet++ as a generator is helpful in enhancing overall needle segmentation performance by correctly recognizing the importance of regions in images. It is also worth noting that the inference time of AdvSeg-Net$^{2}$ is only $6.4\pm2.6~ms$, which can be considered real-time. Including preprocessing and post-processing, the total time is approximately $50~ms$, which is sufficient for handling the US image segmentation problem.

\subsection{Robotic US Probe Repositioning Performance}
\par
To validate the accuracy of the probe repositioning process, we initially identified the best pose for visualizing the inserted needle in an ex vivo porcine sample, namely in-plane pose. Then, we rotate the probe along its centerline in different angles ($\Delta \theta = 5^{\circ}$, $10^{\circ}$, and $15^{\circ}$), and translate the probe in the orthogonal direction $\Delta \textbf{P} = 0~mm$, $3~mm$, and $6~mm$. In total, nine combinations of \{$\Delta \theta$, $\Delta \textbf{P}$\} are tested in the experiment. To ensure statistical consistency, five experiments were carried out independently in each setting. The positional error $e_{p}$ and angular error $e_{\theta}$ are summarized in Table~\ref{tab:performance_reposition}.

\par
The smallest $e_{p}$ is $0.81\pm0.19~mm$ when $\Delta \theta = 5^{\circ}$ and $\Delta \textbf{P} = 0~mm$, while the largest one is $3.88\pm0.52~mm$ when $\Delta \theta = 15^{\circ}$ and $\Delta \textbf{P} = 6~mm$. Regarding $e_{\theta}$, the smallest one is $0.76\pm0.68^{\circ}$ when $\Delta \theta = 5^{\circ}$ and $\Delta \textbf{P} = 0~mm$. The largest $e_{\theta}$ is $1.78\pm0.68$ when $\Delta \theta = 15^{\circ}$ and $\Delta \textbf{P} = 3~mm$. Overall, when the deviation from the ideal pose is small, the precision of probe repositioning is relatively higher. But, it is worth noting that both $e_{p}$ and $e_{\theta}$ are very close in varying initial settings. The overall error across all experimental settings are $1.51\pm0.95~mm$ and $1.25\pm0.79^{\circ}$. In all $45$ experiments, the proposed method can successfully restore the visibility of the inserted needle by maneuvering the US probe.

\section{Conclusion}
\par
This study presents a robotic US system with the capability to monitor the insertion process online and autonomously restore the visibility of the inserted instrument in case of misalignment. The AdvSeg-Net is structured in an adversarial architecture, with the generator predicting the needle from US images and the discriminator ensuring high-order consistencies between the ground truth and the segmented masks. Based on the experiments carried out on ex vivo samples of porcine tissues, we systematically investigated the impacts on segmentation performance using various losses (CE, FL, Dice, and their combinations) and different backbones (UNet, Attention UNet, UNet++, and DeeplabV3+). Taking all factors into consideration, in the final model, we employed a hybrid loss \{Dice, CL, Adv\} as tainting loss and UNet++ as the backbone of the generator. This combination yields the most robust performance in both segmentation-wise and application-wise metrics. Furthermore, the grad-CAM of latent features obtained in various training phases demonstrates that the AdvSeg-Net$^{2}$ is capable of accurately inferring the importance map in images. Leveraging the real-time segmentation results, an online misalignment detection module is developed to track the needle insertion procedure, and a corresponding robotic probe repositioning procedure is carried out to restore the visibility of the needle. In this article, we first highlight the slice thickness artifacts in needle detection, and a correction method is presented to enhance the precision of needle appearance in transverse scanning. The needle appearance is successfully restored under the repositioned probe pose in all 45 trials on unseen ex vivo porcine tissue, with repositioning errors of $1.51\pm0.95mm$ and $1.25\pm0.79^{\circ}$.

\par
In addition to the promising conclusions discussed earlier, there are some noteworthy findings and further perspectives that we want to discuss, hoping they can inspire future advances in this direction. Firstly, instead of adjusting the probe pose, it will also make sense to steer the needle directly for some applications that strictly require a target position, such as tumor ablation. Combining the technologies of steerable needles~\cite{thamo2023towards, sikorski2021flexible} and the side achievement of this study regarding the computation of rotational and positional deviation between the current needle pose and the desired needle pose are promising to enhance the system's adaptability in various clinical scenarios. Secondly, image intensity-based segmentation methods are facing difficulties in terms of US image quality and accurate labeling of thin objects, in particular for 2D US. For example, experienced sonographers sometimes cannot accurately and certainly determine the tip of the needle only from B-Mode images because it is easy to deviate from the probe in-plane during the needle insertion. The new trend of developing native 3D US technologies~\cite{yang2021efficient} is promising to ensure the covering of the percutaneous needle. Thirdly, additional information like vibration~\cite{beigi2017casper}, acoustic physics~\cite{jing2008optimum, palisetti2023acoustic} can be used to further compensate for the intensity-based segmentation method by introducing the new information, enhancing the robustness and precision of needle detection in complex scenarios. Although the method has only been validated on steel needles, it should also be extendable to other types of needles, such as polyethylene, polyurethane, or gallium. However, to achieve optimal segmentation performance, additional datasets should be collected for each type of needle individually.




\section*{ACKNOWLEDGMENT}
This study was supported in part by Brainlab, Germany, and Multi-scale Medical Robotics Center, AIR@InnoHK, Hong Kong. This article reflects the authors’ opinions and conclusions, and not any other entity.


\bibliographystyle{IEEEtran}
\bibliography{IEEEabrv,references}

\begin{thebibliography}{10}
\providecommand{\url}[1]{#1}
\csname url@samestyle\endcsname
\providecommand{\newblock}{\relax}
\providecommand{\bibinfo}[2]{#2}
\providecommand{\BIBentrySTDinterwordspacing}{\spaceskip=0pt\relax}
\providecommand{\BIBentryALTinterwordstretchfactor}{4}
\providecommand{\BIBentryALTinterwordspacing}{\spaceskip=\fontdimen2\font plus
\BIBentryALTinterwordstretchfactor\fontdimen3\font minus \fontdimen4\font\relax}
\providecommand{\BIBforeignlanguage}[2]{{%
\expandafter\ifx\csname l@#1\endcsname\relax
\typeout{** WARNING: IEEEtran.bst: No hyphenation pattern has been}%
\typeout{** loaded for the language `#1'. Using the pattern for}%
\typeout{** the default language instead.}%
\else
\language=\csname l@#1\endcsname
\fi
#2}}
\providecommand{\BIBdecl}{\relax}
\BIBdecl

\bibitem{masoumi2023big}
N.~Masoumi, H.~Rivaz, I.~Hacihaliloglu, M.~O. Ahmad, I.~Reinertsen, and Y.~Xiao, ``The big bang of deep learning in ultrasound-guided surgery: a review,'' \emph{IEEE Transactions on Ultrasonics, Ferroelectrics, and Frequency Control}, 2023.

\bibitem{yang2023medical}
H.~Yang, C.~Shan, A.~F. Kolen, and P.~H. de~With, ``Medical instrument detection in ultrasound: a review,'' \emph{Artificial Intelligence Review}, vol.~56, no.~5, pp. 4363--4402, 2023.

\bibitem{jiang2022precise}
Z.~Jiang, N.~Danis, Y.~Bi, M.~Zhou, M.~Kroenke, T.~Wendler, and N.~Navab, ``Precise repositioning of robotic ultrasound: Improving registration-based motion compensation using ultrasound confidence optimization,'' \emph{IEEE Transactions on Instrumentation and Measurement}, vol.~71, pp. 1--11, 2022.

\bibitem{ottacher2020positional}
D.~Ottacher, A.~Chan, E.~Parent, and E.~Lou, ``Positional and orientational accuracy of 3-d ultrasound navigation system on vertebral phantom study,'' \emph{IEEE Transactions on Instrumentation and Measurement}, vol.~69, no.~9, pp. 6412--6419, 2020.

\bibitem{yang2021automatic}
C.~Yang, M.~Jiang, M.~Chen, M.~Fu, J.~Li, and Q.~Huang, ``Automatic 3-d imaging and measurement of human spines with a robotic ultrasound system,'' \emph{IEEE Transactions on Instrumentation and Measurement}, vol.~70, pp. 1--13, 2021.

\bibitem{li2021overview}
K.~Li, Y.~Xu, and M.~Q.-H. Meng, ``An overview of systems and techniques for autonomous robotic ultrasound acquisitions,'' \emph{IEEE Transactions on Medical Robotics and Bionics}, vol.~3, no.~2, pp. 510--524, 2021.

\bibitem{jiang2023intelligent}
Z.~Jiang, Y.~Bi, M.~Zhou, Y.~Hu, M.~Burke, and N.~Navab, ``Intelligent robotic sonographer: Mutual information-based disentangled reward learning from few demonstrations,'' \emph{The International Journal of Robotics Research}, p. 02783649231223547.

\bibitem{li2024fully}
G.~Li, Q.~Huang, C.~Liu, G.~Wang, L.~Guo, R.~Liu, and L.~Liu, ``Fully automated diagnosis of thyroid nodule ultrasound using brain-inspired inference,'' \emph{Neurocomputing}, vol. 582, p. 127497, 2024.

\bibitem{huang2024robot}
Q.~Huang, B.~Gao, and M.~Wang, ``Robot-assisted autonomous ultrasound imaging for carotid artery,'' \emph{IEEE Transactions on Instrumentation and Measurement}, 2024.

\bibitem{huang2023mimicking}
Q.~Huang, Y.~Wang, H.~Luo, and J.~Li, ``On mimicking human’s manipulation for robot-assisted spine ultrasound imaging,'' \emph{Robotic Intelligence and Automation}, vol.~43, no.~4, pp. 373--381, 2023.

\bibitem{bi2024machine}
Y.~Bi, Z.~Jiang, F.~Duelmer, D.~Huang, and N.~Navab, ``Machine learning in robotic ultrasound imaging: Challenges and perspectives,'' \emph{Annual Review of Control, Robotics, and Autonomous Systems}, vol.~7.

\bibitem{jiang2022towards}
Z.~Jiang, Y.~Gao, L.~Xie, and N.~Navab, ``Towards autonomous atlas-based ultrasound acquisitions in presence of articulated motion,'' \emph{IEEE Robotics and Automation Letters}, vol.~7, no.~3, pp. 7423--7430, 2022.

\bibitem{zhao2019electromagnetic}
Z.~Zhao and Z.~T.~H. Tse, ``An electromagnetic tracking needle clip: an enabling design for low-cost image-guided therapy,'' \emph{Minimally Invasive Therapy \& Allied Technologies}, vol.~28, no.~3, pp. 165--171, 2019.

\bibitem{novotny2003tool}
P.~M. Novotny, J.~W. Cannon, and R.~D. Howe, ``Tool localization in 3d ultrasound images,'' in \emph{Medical Image Computing and Computer-Assisted Intervention-MICCAI 2003: 6th International Conference, Montr{\'e}al, Canada, November 15-18, 2003. Proceedings 6}.\hskip 1em plus 0.5em minus 0.4em\relax Springer, 2003, pp. 969--970.

\bibitem{beigi2021enhancement}
P.~Beigi, S.~E. Salcudean, G.~C. Ng, and R.~Rohling, ``Enhancement of needle visualization and localization in ultrasound,'' \emph{International journal of computer assisted radiology and surgery}, vol.~16, pp. 169--178, 2021.

\bibitem{kaya2014needle}
M.~Kaya and O.~Bebek, ``Needle localization using gabor filtering in 2d ultrasound images,'' in \emph{2014 IEEE International Conference on Robotics and Automation (ICRA)}.\hskip 1em plus 0.5em minus 0.4em\relax IEEE, 2014, pp. 4881--4886.

\bibitem{litjens2017survey}
G.~Litjens, T.~Kooi, B.~E. Bejnordi, A.~A.~A. Setio, F.~Ciompi, M.~Ghafoorian, J.~A. Van Der~Laak, B.~Van~Ginneken, and C.~I. S{\'a}nchez, ``A survey on deep learning in medical image analysis,'' \emph{Medical image analysis}, vol.~42, pp. 60--88, 2017.

\bibitem{jiang2024class}
Z.~Jiang, Y.~Kang, Y.~Bi, X.~Li, C.~Li, and N.~Navab, ``Class-aware cartilage segmentation for autonomous us-ct registration in robotic intercostal ultrasound imaging,'' \emph{IEEE Transactions on Automation Science and Engineering}, 2024.

\bibitem{bi2023mi}
Y.~Bi, Z.~Jiang, R.~Clarenbach, R.~Ghotbi, A.~Karlas, and N.~Navab, ``Mi-segnet: Mutual information-based us segmentation for unseen domain generalization,'' in \emph{International Conference on Medical Image Computing and Computer-Assisted Intervention}.\hskip 1em plus 0.5em minus 0.4em\relax Springer, 2023, pp. 130--140.

\bibitem{gillies2020deep}
D.~J. Gillies, J.~R. Rodgers, I.~Gyacskov, P.~Roy, N.~Kakani, D.~W. Cool, and A.~Fenster, ``Deep learning segmentation of general interventional tools in two-dimensional ultrasound images,'' \emph{Medical Physics}, vol.~47, no.~10, pp. 4956--4970, 2020.

\bibitem{ronneberger2015u}
O.~Ronneberger, P.~Fischer, and T.~Brox, ``U-net: Convolutional networks for biomedical image segmentation,'' in \emph{International Conference on Medical image computing and computer-assisted intervention}.\hskip 1em plus 0.5em minus 0.4em\relax Springer, 2015, pp. 234--241.

\bibitem{chen2022automatic}
S.~Chen, Y.~Lin, Z.~Li, F.~Wang, and Q.~Cao, ``Automatic and accurate needle detection in 2d ultrasound during robot-assisted needle insertion process,'' \emph{International Journal of Computer Assisted Radiology and Surgery}, pp. 1--9, 2022.

\bibitem{lee2020ultrasound}
J.~Y. Lee, M.~Islam, J.~R. Woh, T.~M. Washeem, L.~Y.~C. Ngoh, W.~K. Wong, and H.~Ren, ``Ultrasound needle segmentation and trajectory prediction using excitation network,'' \emph{International Journal of Computer Assisted Radiology and Surgery}, vol.~15, no.~3, pp. 437--443, 2020.

\bibitem{chaurasia2017linknet}
A.~Chaurasia and E.~Culurciello, ``Linknet: Exploiting encoder representations for efficient semantic segmentation,'' in \emph{2017 IEEE visual communications and image processing (VCIP)}.\hskip 1em plus 0.5em minus 0.4em\relax IEEE, 2017, pp. 1--4.

\bibitem{mwikirize2018convolution}
C.~Mwikirize, J.~L. Nosher, and I.~Hacihaliloglu, ``Convolution neural networks for real-time needle detection and localization in 2d ultrasound,'' \emph{International journal of computer assisted radiology and surgery}, vol.~13, pp. 647--657, 2018.

\bibitem{hacihaliloglu2015projection}
I.~Hacihaliloglu, P.~Beigi, G.~Ng, R.~N. Rohling, S.~Salcudean, and P.~Abolmaesumi, ``Projection-based phase features for localization of a needle tip in 2d curvilinear ultrasound,'' in \emph{Medical Image Computing and Computer-Assisted Intervention--MICCAI 2015: 18th International Conference, Munich, Germany, October 5-9, 2015, Proceedings, Part I 18}.\hskip 1em plus 0.5em minus 0.4em\relax Springer, 2015, pp. 347--354.

\bibitem{mwikirize2021time}
C.~Mwikirize, A.~B. Kimbowa, S.~Imanirakiza, A.~Katumba, J.~L. Nosher, and I.~Hacihaliloglu, ``Time-aware deep neural networks for needle tip localization in 2d ultrasound,'' \emph{International Journal of Computer Assisted Radiology and Surgery}, vol.~16, pp. 819--827, 2021.

\bibitem{beigi2017casper}
P.~Beigi, R.~Rohling, S.~E. Salcudean, and G.~C. Ng, ``Casper: computer-aided segmentation of imperceptible motion—a learning-based tracking of an invisible needle in ultrasound,'' \emph{International journal of computer assisted radiology and surgery}, vol.~12, pp. 1857--1866, 2017.

\bibitem{yan2023learning}
W.~Yan, Q.~Ding, J.~Chen, K.~Yan, R.~S.-Y. Tang, and S.~S. Cheng, ``Learning-based needle tip tracking in 2d ultrasound by fusing visual tracking and motion prediction,'' \emph{Medical Image Analysis}, vol.~88, p. 102847, 2023.

\bibitem{zhang2020multi}
Y.~Zhang, X.~He, Z.~Tian, J.~J. Jeong, Y.~Lei, T.~Wang, Q.~Zeng, A.~B. Jani, W.~J. Curran, P.~Patel \emph{et~al.}, ``Multi-needle detection in 3d ultrasound images using unsupervised order-graph regularized sparse dictionary learning,'' \emph{IEEE transactions on medical imaging}, vol.~39, no.~7, pp. 2302--2315, 2020.

\bibitem{arif2019automatic}
M.~Arif, A.~Moelker, and T.~van Walsum, ``Automatic needle detection and real-time bi-planar needle visualization during 3d ultrasound scanning of the liver,'' \emph{Medical image analysis}, vol.~53, pp. 104--110, 2019.

\bibitem{yang2020efficient}
H.~Yang, C.~Shan, A.~F. Kolen, and P.~H. de~With, ``Efficient medical instrument detection in 3d volumetric ultrasound data,'' \emph{IEEE Transactions on Biomedical Engineering}, vol.~68, no.~3, pp. 1034--1043, 2020.

\bibitem{yang2021efficient}
H.~Yang, C.~Shan, A.~Bouwman, A.~F. Kolen, and P.~H. de~With, ``Efficient and robust instrument segmentation in 3d ultrasound using patch-of-interest-fusenet with hybrid loss,'' \emph{Medical Image Analysis}, vol.~67, p. 101842, 2021.

\bibitem{luc2016semantic}
P.~Luc, C.~Couprie, S.~Chintala, and J.~Verbeek, ``Semantic segmentation using adversarial networks,'' \emph{arXiv preprint arXiv:1611.08408}, 2016.

\bibitem{oktay2022attention}
O.~Oktay, J.~Schlemper, L.~Le~Folgoc, M.~Lee, M.~Heinrich, K.~Misawa, K.~Mori, S.~McDonagh, N.~Y. Hammerla, B.~Kainz \emph{et~al.}, ``Attention u-net: Learning where to look for the pancreas,'' in \emph{Medical Imaging with Deep Learning}, 2022.

\bibitem{zhou2018unet++}
Z.~Zhou, M.~M. Rahman~Siddiquee, N.~Tajbakhsh, and J.~Liang, ``Unet++: A nested u-net architecture for medical image segmentation,'' in \emph{Deep Learning in Medical Image Analysis and Multimodal Learning for Clinical Decision Support: 4th International Workshop, DLMIA 2018, and 8th International Workshop, ML-CDS 2018, Held in Conjunction with MICCAI 2018, Granada, Spain, September 20, 2018, Proceedings 4}.\hskip 1em plus 0.5em minus 0.4em\relax Springer, 2018, pp. 3--11.

\bibitem{chen2018encoder}
L.-C. Chen, Y.~Zhu, G.~Papandreou, F.~Schroff, and H.~Adam, ``Encoder-decoder with atrous separable convolution for semantic image segmentation,'' in \emph{Proceedings of the European conference on computer vision (ECCV)}, 2018, pp. 801--818.

\bibitem{radford2015unsupervised}
A.~Radford, L.~Metz, and S.~Chintala, ``Unsupervised representation learning with deep convolutional generative adversarial networks,'' \emph{arXiv preprint arXiv:1511.06434}, 2015.

\bibitem{bertels2019optimizing}
J.~Bertels, T.~Eelbode, M.~Berman, D.~Vandermeulen, F.~Maes, R.~Bisschops, and M.~B. Blaschko, ``Optimizing the dice score and jaccard index for medical image segmentation: Theory and practice,'' in \emph{International conference on medical image computing and computer-assisted intervention}.\hskip 1em plus 0.5em minus 0.4em\relax Springer, 2019, pp. 92--100.

\bibitem{lin2017focal}
T.-Y. Lin, P.~Goyal, R.~Girshick, K.~He, and P.~Doll{\'a}r, ``Focal loss for dense object detection,'' in \emph{Proceedings of the IEEE international conference on computer vision}, 2017, pp. 2980--2988.

\bibitem{milletari2016v}
F.~Milletari, N.~Navab, and S.-A. Ahmadi, ``V-net: Fully convolutional neural networks for volumetric medical image segmentation,'' in \emph{2016 fourth international conference on 3D vision (3DV)}.\hskip 1em plus 0.5em minus 0.4em\relax Ieee, 2016, pp. 565--571.

\bibitem{selvaraju2017grad}
R.~R. Selvaraju, M.~Cogswell, A.~Das, R.~Vedantam, D.~Parikh, and D.~Batra, ``Grad-cam: Visual explanations from deep networks via gradient-based localization,'' in \emph{Proceedings of the IEEE international conference on computer vision}, 2017, pp. 618--626.

\bibitem{alsinan2020bone}
A.~Z. Alsinan, V.~M. Patel, and I.~Hacihaliloglu, ``Bone shadow segmentation from ultrasound data for orthopedic surgery using gan,'' \emph{International Journal of Computer Assisted Radiology and Surgery}, vol.~15, no.~9, pp. 1477--1485, 2020.

\bibitem{mechrez2018contextual}
R.~Mechrez, I.~Talmi, and L.~Zelnik-Manor, ``The contextual loss for image transformation with non-aligned data,'' in \emph{Proceedings of the European conference on computer vision (ECCV)}, 2018, pp. 768--783.

\bibitem{simonyan2014very}
K.~Simonyan and A.~Zisserman, ``Very deep convolutional networks for large-scale image recognition,'' \emph{arXiv preprint arXiv:1409.1556}, 2014.

\bibitem{deng2009imagenet}
J.~Deng, W.~Dong, R.~Socher, L.-J. Li, K.~Li, and L.~Fei-Fei, ``Imagenet: A large-scale hierarchical image database,'' in \emph{2009 IEEE conference on computer vision and pattern recognition}.\hskip 1em plus 0.5em minus 0.4em\relax Ieee, 2009, pp. 248--255.

\bibitem{han2018spine}
Z.~Han, B.~Wei, A.~Mercado, S.~Leung, and S.~Li, ``Spine-gan: Semantic segmentation of multiple spinal structures,'' \emph{Medical image analysis}, vol.~50, pp. 23--35, 2018.

\bibitem{shorten2019survey}
C.~Shorten and T.~M. Khoshgoftaar, ``A survey on image data augmentation for deep learning,'' \emph{Journal of big data}, vol.~6, no.~1, pp. 1--48, 2019.

\bibitem{kingma2014adam}
D.~P. Kingma and J.~Ba, ``Adam: A method for stochastic optimization,'' in \emph{International Conference on Learning Representations (ICLR)}, 2015.

\bibitem{goldstein1981slice}
A.~Goldstein and B.~L. Madrazo, ``Slice-thickness artifacts in gray-scale ultrasound,'' \emph{Journal of clinical ultrasound}, vol.~9, no.~7, pp. 365--375, 1981.

\bibitem{peikari2012characterization}
M.~Peikari, T.~K. Chen, A.~Lasso, T.~Heffter, G.~Fichtinger, and E.~C. Burdette, ``Characterization of ultrasound elevation beamwidth artifacts for prostate brachytherapy needle insertion,'' \emph{Medical physics}, vol.~39, no.~1, pp. 246--256, 2012.

\bibitem{peikari2011effects}
M.~Peikari, T.~K. Chen, A.~Lasso, T.~Heffter, and G.~Fichtinger, ``Effects of ultrasound section-thickness on brachytherapy needle tip localization error,'' in \emph{Medical Image Computing and Computer-Assisted Intervention--MICCAI 2011: 14th International Conference, Toronto, Canada, September 18-22, 2011, Proceedings, Part I 14}.\hskip 1em plus 0.5em minus 0.4em\relax Springer, 2011, pp. 299--306.

\bibitem{jiang2021autonomous}
Z.~Jiang, Z.~Li, M.~Grimm, M.~Zhou, M.~Esposito, W.~Wein, W.~Stechele, T.~Wendler, and N.~Navab, ``Autonomous robotic screening of tubular structures based only on real-time ultrasound imaging feedback,'' \emph{IEEE Transactions on Industrial Electronics}, 2021.

\bibitem{jiang2023robotic}
Z.~Jiang, S.~E. Salcudean, and N.~Navab, ``Robotic ultrasound imaging: State-of-the-art and future perspectives,'' \emph{Medical image analysis}, p. 102878, 2023.

\bibitem{ester1996density}
M.~Ester, H.-P. Kriegel, J.~Sander, X.~Xu \emph{et~al.}, ``A density-based algorithm for discovering clusters in large spatial databases with noise,'' in \emph{kdd}, vol.~96, no.~34, 1996, pp. 226--231.

\bibitem{Mariga_pyRANSAC-3D_2022}
\BIBentryALTinterwordspacing
L.~Mariga, ``{pyRANSAC-3D},'' 10 2022. [Online]. Available: \url{https://github.com/leomariga/pyRANSAC-3D}
\BIBentrySTDinterwordspacing

\bibitem{hennersperger2016towards}
C.~Hennersperger, B.~Fuerst, and et~al., ``Towards {MRI}-based autonomous robotic us acquisitions: a first feasibility study,'' \emph{IEEE Trans. Med. Imaging}, vol.~36, no.~2, pp. 538--548, 2016.

\bibitem{thamo2023towards}
B.~Thamo, V.~Voulgaridou, H.~Wood, J.~Stone, K.~Dhaliwal, and M.~Khadem, ``Towards robotics-assisted endomicroscopy in percutaneous needle-based interventions,'' \emph{IEEE Transactions on Medical Robotics and Bionics}, 2023.

\bibitem{sikorski2021flexible}
J.~Sikorski, C.~M. Heunis, R.~Obeid, V.~K. Venkiteswaran, and S.~Misra, ``A flexible catheter system for ultrasound-guided magnetic projectile delivery,'' \emph{IEEE transactions on robotics}, vol.~38, no.~3, pp. 1959--1972, 2021.

\bibitem{jing2008optimum}
Y.~Jing, R.~Bocala, A.~Oberai, and P.~Bigeleisen, ``Optimum design of echogenic needles for ultrasound guided nerve block,'' in \emph{2008 IEEE Ultrasonics Symposium}.\hskip 1em plus 0.5em minus 0.4em\relax IEEE, 2008, pp. 1334--1337.

\bibitem{palisetti2023acoustic}
A.~Palisetti, A.~King, C.~Priestner, H.~Yi, Y.~Tang, R.~Murakami, and H.~K. Zhang, ``Acoustic reflector-integrated encapsulation for needle-aligned ultrasound-guided pcnl access,'' in \emph{2023 IEEE International Ultrasonics Symposium (IUS)}.\hskip 1em plus 0.5em minus 0.4em\relax IEEE, 2023, pp. 1--4.

\end{thebibliography}

\end{document}